# Deep learning for conifer/deciduous classification of airborne LiDAR 3D point clouds representing individual trees

Hamid Hamraz[a*], Nathan B. Jacobs[a], Marco A. Contreras[b], and Chase H. Clark[b]

a: Department of Computer Science, b: Department of Forestry

University of Kentucky, Lexington, KY 40506, USA

[hhamraz, jacobs] @cs.uky.edu, [marco.contreras, chase.clark2] @uky.edu

\* Corresponding Author:   hhamraz@cs.uky.edu   +1 (859) 489 1261

**Abstract.** The purpose of this study was to investigate the use of deep learning for coniferous/deciduous classification of individual trees from airborne LiDAR data. To enable efficient processing by a deep convolutional neural network (CNN), we designed two discrete representations using leaf-off and leaf-on LiDAR data: a digital surface model with four channels (DSM×4) and a set of four 2D views (4×2D). A training dataset of labeled tree crowns was generated via segmentation of tree crowns, followed by co-registration with field data. Potential mislabels due to GPS error or tree leaning were corrected using a statistical ensemble filtering procedure. Because the training data was heavily unbalanced (~8% conifers), we trained an ensemble of CNNs on random balanced sub-samples of augmented data (180 rotational variations per instance). The 4×2D representation yielded similar classification accuracies to the DSM×4 representation (~82% coniferous and ~90% deciduous) while converging faster. The data augmentation improved the classification accuracies, but more real training instances (especially coniferous) likely results in much stronger improvements. Leaf-off LiDAR data were the primary source of useful information, which is likely due to the perennial nature of coniferous foliage. LiDAR intensity values also proved to be useful, but normalization yielded no significant improvements. As we observed, large training data may compensate for the lack of a subset of important domain data. Lastly, the classification accuracies of overstory trees (~90%) were more balanced than those of understory trees (~90% deciduous and ~65% coniferous), which is likely due to the incomplete capture of understory tree crowns via airborne LiDAR. Automatic derivation of optimal features via deep learning provide the opportunity for remarkable improvements in prediction tasks where captured data are not friendly to human visual system – likely yielding sub-optimal human-designed features.

# 1 Introduction

Remote sensing technologies have long been a means to facilitate data acquisition over large forested areas (Franklin 2001). For instance, aerial images have been used to map forests and monitor their growth and regeneration (Gougeon 1995; Pitkänen 2001; Quackenbush et al. 2000). However, 2D images, as snapshots of the 3D world, lose depth information and are insufficient for more detailed estimation tasks such as the derivation of vertical canopy structure and biomass quantification. Airborne light detection and ranging (LiDAR) directly measures depth and can capture multiple returns per emission, thereby representing the forested landscapes in the form of 3D point clouds (Ackermann 1999; Hyyppä et al. 2012; Maltamo et al. 2014). These point clouds can be processed to segment individual trees (Amiri et al. 2016; Jing et al. 2012; Kwak et al. 2007; Paris et al. 2016; Popescu and Zhao 2008; Sačkov et al. 2017; Véga et al. 2014; Wang et al. 2008), which can then be used to derive tree allometric dimensions such as height and crown width, and to predict different tree attributes such as type, species, status (live or dead), or diameter at breast height (DBH) (Duncanson et al. 2015; Vauhkonen et al. 2010; Yu et al. 2011).

Several studies have used segmented point clouds representing individual trees in an attempt to predict tree type (coniferous or deciduous) or species using machine learning methods (Blomley et al. 2017; Cao et al. 2016; Harikumar et al. 2017; Holmgren and Persson 2004; Kim et al. 2011; Lindberg et al. 2014; Ørka et al. 2009; Reitberger et al. 2008). In these studies, researchers derived a set of features related to crown geometry and foliage density/pattern/texture from the LiDAR data and input the features into different classification methods such as linear discriminant analysis, k-nearest neighbors, random forest, and support vector machines (SVMs). A few studies have presented automated or semi-automated approaches for identifying useful features for the task of tree species classification (Bruggisser et al. 2017; Li et al. 2013; Lin and Hyyppä 2016). The previous work using traditional learning methods has required that the set of candidate features be assembled by an expert, with the intention of removing redundant and less useful information from the raw data. However, because LiDAR point clouds are not easily processed by the human visual system, the expert-designed features may as well be suboptimal and likely missing useful information.

Deep neural network learning methods, on the other hand, can directly map the input raw data to the target prediction by passing the input through multiple layers (LeCun et al. 2015; Schmidhuber 2015). The initial layers are designed to extract the useful low to high level features, and the next layers map the extracted features to the target prediction. The advantage of deep learning methods is in their end-to-end operation, i.e., both feature extraction and mapping to the target prediction are trained automatically as a whole such that the global prediction task functions optimally. Although some expertise is required to set up a reasonable deep network architecture and tune the optimization hyper-parameters, no human intervention is required for feature extraction.

A large body of research has been devoted to a variety of deep learning classification or segmentation tasks using 2D images as the raw input data (Girshick et al. 2014; Krizhevsky et al. 2012). However, 3D data have attracted less attention due to more costly acquisition/processing and their less intuitive and less conventional representational formats, which demand non-trivial pre-processing techniques to discretize the data and make them usable for deep learning methods (Qi et al. 2017; Qi et al. 2016). A number of studies have binned 3D data into voxel spaces to create representations that can be input to and processed by a 3D convolutional neural network (CNN) (Dou et al. 2016; Maturana and Scherer 2015; Wu et al. 2015). Although voxel spaces are perhaps the most comprehensive discrete representations that preserve the raw 3D structure, they are computationally expensive to process, more prone to overfitting, and therefore prohibitive for use with larger datasets. Other studies have created 2.5D digital surface models (DSMs) (Mizoguchi et al. 2017; Roth et al. 2016; Socher et al. 2012) or multiple 2D views (Farfade et al. 2015; Su et al. 2015) from the 3D data. In the event that 3D imaging/sensing technology can capture the internal structure of the measured objects, conversion to DSM or 2D views may forego this internal structure. However, depending on the application, DSMs and/or multiple 2D views can provide as much useful information as a full 3D representation while being less prone to overfitting and incurring less computational cost (Kalogerakis et al. 2017; Su et al. 2015).

A few recent studies used deep learning methods to classify species of individual trees from very high resolution ground-based LiDAR point clouds. Guan et al. (2015) segmented individual trees from mobile LiDAR point clouds in an urban area, developed a waveform representation to model the geometry of the trees, and used deep learning to convert the waveform representation to high-level features. These features were then input to an SVM classifier to perform tree species classification. Mizoguchi et al. (2017) also segmented individual trees from terrestrial LiDAR point clouds, derived DSM patches representing the tree bark texture from the clouds, and fed this information into a CNN to perform classification between two species. In contrast to ground-based LiDAR, airborne LiDAR provides information over a much larger scale and from an entirely different perspective. However, we could not identify any deep learning studies concerned with individual tree classification from airborne LiDAR data.

In this paper, we segment individual trees from airborne LiDAR data representing a natural dense forest, discretize segmented crowns to input them into a CNN, and perform different deep learning experiments on tree type classification. The main contributions of this work are: (i) to engineer two discrete formats for the 3D crown cloud captured by airborne LiDAR and thereby enable efficient processing by a CNN architecture, and (ii) to investigate the effects of different design decisions with respect to training data preparation and deep network structure as well as the effects of training data composition and domain-specific data on the classification accuracy.

## 2 Materials and Methods

### 2.1 Study site, LiDAR campaign, and field survey

The study site is the University of Kentucky's Robinson Forest (RF, Lat. 37.4611, Long. -83.1555) which is located in the rugged eastern section of the Cumberland Plateau region of southeastern Kentucky in Breathitt, Perry, and Knott counties. RF features a variable, dissected topography with moderately steep slopes, which range from 10% to over 100% and face predominately northwest to southeast. Elevation ranges from 252 to 503 meters above sea level (Carpenter and Rumsey 1976). RF is composed of a diverse, contiguous, mixed mesophytic vegetation made up of various deciduous tree species with northern red oak (*Quercus rubra*), white oak (*Quercus alba*), yellow-poplar (*Liriodendron tulipifera*), American beech (*Fagus grandifolia*), and sugar maple (*Acer saccharum*) as overstory species. Deciduous understory species include eastern redbud (*Cercis canadensis*), flowering dogwood (*Cornus florida*), spicebush (*Lindera benzoin*), pawpaw (*Asimina triloba*), umbrella magnolia *(Magnolia tripetala)*, and bigleaf magnolia (*Magnolia macrophylla*) (Carpenter and Rumsey 1976; Overstreet 1984). Additionally, a small number of conifer species also exists throughout the forest including eastern hemlock (*Tsuga canadensis*), which can occur in clusters near streams, and different sub-species of Pine (*Pinus sp*). Average stem density in RF is about 330 stems per hectare (Hamraz et al. 2017b) and average canopy cover is about 93% with small openings scattered throughout. Most areas exceed 97% canopy cover, but recently harvested areas have an average cover as low as 63%. Having been extensively logged in the 1920's, RF is considered a second growth forest ranging from 80-100 years old, and it is now protected from commercial logging and mining activities (Department of Forestry 2007). RF currently covers an aggregate area of ~7,440 ha and includes about 2.5 million (±5.6%) trees (Hamraz and Contreras 2017).

The LiDAR data are a combination of two separate datasets collected with the same LiDAR system (Leica ALS60). For both datasets, the system was set at 200 kHz pulse repetition rate and 40° field of view, and was flown with an average speed of 105 knots over strips with 50% overlap. One dataset was low density (~2 pt/m²), collected in the spring of 2013 during the leaf-off season (average altitude of 3,096 m above the ground) for the purpose of acquiring terrain information as a part of a state-wide elevation data acquiring program by the Kentucky Division of Geographic Information. The second dataset was high density (~50 pt/m²), collected in the summer of 2013 during the leaf-on season (average altitude of 214 m above the ground). Up to three and four returns were captured per pulse for the leaf-off and the leaf-on collections respectively, and only 90–95% of the middle portion of the flight strips was used to create the datasets. Both datasets were processed by the vendor using TerraScan software (Terrasolid Ltd. 2012). In this process, LiDAR points were classified into ground and non-ground points. Ground points were then used to create a 1-meter resolution digital elevation model (DEM) using nearest neighbor as the fill void method and the average as the interpolation method.

Throughout the entire RF, 271 regularly distributed (grid-wise every 384 m) circular plots of 0.04 ha, centers of which were georeferenced with 5 m accuracy, were field surveyed during the summer of 2013. Within each plot, DBH (cm), tree height (m), species, crown class (dominant, co-dominant, intermediate, overtopped), tree status (live, dead), and stem class (single, multiple) were recorded for all trees with DBH > 12.5 cm. In addition, horizontal distance and azimuth from plot center to the face of each tree at breast height were collected to create a stem map. Excluding trees below 4 m in height, a total of 3987 trees were surveyed of which 7.27% were conifers (Table 1).

Table 1. Summary statistics of trees surveyed within 271 plots in Robinson Forest.

|  | Conifer | Percent in Conifers | Deciduous | Percent in Deciduous | Total | Percent in Total |
|---|---|---|---|---|---|---|
| **Dominant** | 10 | 3.45% | 120 | 3.46% | 130 | 3.26% |
| **Co-Dominant** | 39 | 13.45% | 919 | 24.86% | 958 | 24.03% |
| **Intermediate** | 78 | 26.90% | 1409 | 38.12% | 1487 | 37.30% |
| **Overtopped** | 143 | 49.3% | 1012 | 27.38% | 1155 | 28.97% |
| **Dead** | 20 | 6.90% | 236 | 6.39% | 256 | 6.42% |
| **All** | 290 | 100.0% | 3697 | 100.0% | 3987 | 100.0% |
| **Percent of Total** | 7.27% |  | 92.73% |  | 100.0% |  |
| **Species Count** | 6 |  | 37 |  |  |  |
| **Shannon Diversity Index** | 0.605 |  | 2.673 |  |  |  |

## 2.2 Data preparation

### 2.2.1 LiDAR intensity normalization

The LiDAR intensity value that is recorded for each return is dependent on various factors, many of which are unrelated to the vegetation texture (Gatziolis 2011; Kashani et al. 2015). The distance a LiDAR pulse travels (referred to as range), the angle at which the pulse is emitted, and the LiDAR return number are among the factors affecting intensity that can be controlled for, while different atmospheric factors are difficult to track. Assuming constant atmospheric conditions, we used a data-driven approach to normalize the intensity values. We binned the entire forest dataset to a horizontal grid with a cell width of 10 m and randomly sampled one leaf-off and one leaf-on vegetation point per grid cell. We then grouped the leaf-off and the leaf-on samples according to the return number to create three leaf-off and four leaf-on datasets. For each of the seven datasets, we built a regression model that predicted intensity based on range and emission angle. For the leaf-on datasets, the effect of range and angle was significant: the natural logarithm of range had a negative correlation with intensity ($P < .0001$), and the cosine of angle has a positive correlation (P < .0001) with intensity. However, we did not observe any significant correlations between range/angle and intensity for the leaf-off datasets, which is likely due to the recording of very low intensity values and discretization to an eight-bit format, dimming away such correlations. For each of the four leaf-on datasets, we removed the effects of range and emission angle by residualization (Allen 1997), i.e., we replaced the intensity values by the corresponding model residuals. Finally, we normalized the residualized intensities back to an eight-bit format.

### 2.2.2 Individual tree segmentation and registration with field data

We included the points within a 10 m buffer around the LiDAR point clouds corresponding to the 271 field-surveyed plots in order to capture the complete crowns of border trees. Using the DEM, we calculated the height above ground for the LiDAR points and excluded the points below 3 m (ground level vegetation). We then vertically stratified the point clouds into multiple canopy layers by analyzing the vertical distributions of the LiDAR points within overlapping locales (Hamraz et al. 2017c). We excluded the canopy layers with densities less than 3 pt/m² from further analysis because tree segmentation at such low densities becomes inaccurate (Evans et al. 2009; Hamraz et al. 2017a). We then segmented each of the canopy layers independently using the method we designed for complex vegetation structures (Hamraz et al. 2016). This method identifies crown boundaries around the global maximum of the canopy layer and clusters the points encompassed by the convex hull of boundary points to complete the segmentation for the tallest tree. This process is repeated until all of the points are clustered. Clusters representing crowns less than 1.5 m in average width are finally removed as noise.

In order to register the segmented crowns with the field data, we assigned a score to each pair of segmented crown and field-measured stem locations. The location of each segmented crown was taken from the crown apex. Scores were assigned based on the difference in tree height and the leaning angle from nadir between the crown apex and the stem location. If the height difference was less than 10% and the leaning angle was less than 5°, a score of 100 was assigned. If the height difference and leaning angle were less than 20% and 10° respectively, a score of 70 was assigned. If the height difference and leaning angle were less than 30% and 15°, a score of 40 was assigned. Otherwise, a score of 0 was assigned. We then selected the set of pairs with the maximum total score where each crown or stem location appears not more than once using the Hungarian assignment algorithm and regarded the set as the co-registered tree pairs (Hamraz et al. 2016; Kuhn 1955). Excluding dead trees, a total of 2528 co-registered trees was gleaned, of which 124 (4.90%) were conifers and 2404 (95.10%) were deciduous. Smaller understory trees, especially those represented by very low point densities, were automatically excluded through the segmentation and registration process.

### 2.2.3 Discretization of segmented point clouds

We converted the point cloud of each tree crown to two different representational formats: (1) a DSM with four channels (DSM×4), and (2) a set of four single-channel 2D images (4×2D). To create the DSM×4 format, we binned the point cloud to a horizontal grid of 128×128 pixels of width 12.5 cm such that the apex of the segmented crown would fall in the center pixel. We then recorded the four channel values for each pixel, which included the elevation above ground of the highest leaf-on point, the normalized intensity of the highest leaf-on point, the elevation above ground of the highest leaf-off point, and the intensity for the highest leaf-off point. We chose the small pixel width of 12.5 cm for creating the DSM image to minimize the information loss because of falling multiple LiDAR points in a pixel. The resulting DSM structure captures a square of 16×16 m in the real world, which is large enough to encompass an entire tree crown in almost all cases given that tree crowns are often very tightly situated in dense forests. However, because crown width information may be missing for some large trees, we recorded the crown area as a separate feature alongside the DSM×4 representation.

To create the 4×2D format, we generated one pair of aerial view images and one pair of side profile view images for each segmented crown. One image in each pair was created from the leaf-on point cloud, and the other was created from the leaf-off point cloud. As with the DSM×4 format, the aerial images for a single tree crown covered a square area of 16×16 m, with the crown apex located in the center of the images. The pixel width however, was set to 25 cm because depth information was not intended to be captured in the aerial view. To create the aerial images, we recorded the intensity of the highest LiDAR point in each pixel. The side profile images were created from vertical profiles of the point clouds, which had a thickness of 75 cm and passed through the crown apex. Each of the side view images captured a square area of 16×16 m with a pixel width of 25 cm. The LiDAR point representing the apex was located in the top center pixel. We recorded the mean intensity of leaf-on/leaf-off LiDAR points in the profile for each pixel. Although the majority of trees in our dataset are taller than 16 m, most airborne LiDAR points are recorded in the upper parts of the tree crowns and therefore, a 16 m side view height was deemed sufficient to capture the crown structure that is represented by the LiDAR points. However, because tree height information was missing from both the aerial and side views, we recorded height and crown width as two separate features alongside the 4x2D representation.

The DSM×4 format resembles the 3D point cloud data by losing less 3D structure while the 4×2D format only captures the 3D data from two 2D views taking the advantage of the symmetry of an ideally shaped tree crown. To augment the data and increase the training data size for deep learning experiments, we created the DSM×4 and the 4×2D representational formats over 180 rotational variations of each point cloud. We iteratively rotated the point cloud along a nadir axis through the apex by 2° and created a DSM× 4and a 4×2D representation in each iteration. Although the 4×2D format loses much of the 3D information because in reality the tree crowns has several dissymmetrical structural features, this information has mostly been re-gained when using 180 rotational augmentations per instance.

## 2.3 Convolutional Neural Network models

For the DSM×4 input format, we stacked six pairs of convolutional and max pooling layers including ReLU activation units. Each convolutional layer included one filter of 4×3×3 with a stride of one pixel that was operating on a zero-padded input to maintain the same size for the output. Each max pooling layer included 2×2 max pooling windows per channel, down-sampling the convoluted input to half of the width and the height. Operating on the representation input of size 4×128×128, this layer composition produces a 4×2×2 output structure, which is flattened to 16 output units. On the other hand, for the crown area input feature, we stacked two dense layers, each including two ReLU units. We then put the 16 units initiated from the DSM image and the two units initiated from the crown area feature together and stacked two dense layers of 25 and 10 ReLU units respectively to the end. Finally, we added a softmax layer to obtain the probability distribution over one-hot-encoded class labels.

For the 4×2D input format, we stacked five pairs of convolutional and max pooling layers, including ReLU activation units, per each single-channel 2D image. Each convolutional layer included one filter of size 1×3×3×1 with a stride of one pixel that was operating on a zero-padded input. Each max pooling layer included windows of 2×2, down-sampling the convoluted input image to half of the width and the height. Operating on the set of four image representation inputs of size 1×64×64, this layer composition produces a 2×2×4 output structure, which is flattened to 16 output units. On the other hand, for the crown width and the tree height input features, we stacked two dense layers, including four and two ReLU units respectively. Similar to the DSM network, we put the previous 18 units together and added two dense layers of 25 and 10 ReLU units and a final softmax layer respectively to the end.

The DSM×4 format allows the deep network architecture to perform an early fusion of the leaf-on and leaf-off data as well as the intensity and height values associated with the data. The network captures the correlation between the four channels for the classification task by including more parameters and intermediate features. On the other hand, the 4×2D format allows a late fusion to the network, i.e., the leaf-off and leaf-on data and their intensity/height values are not fused until after the corresponding convolutional and max pooling layers produced features independently. While the DSM×4 format allows for a richer training model, the 4×2D format incurs less computational cost.

## 2.4 Mislabel correction via iterative resampling

As described earlier, registration of the segmented tree crowns to the field-surveyed tree stem locations was done through a probabilistic scoring process. Moreover, the GPS error for the field-surveyed plot centers (~5 m) can exceed the distances between individual trees. These issues likely resulted in a fraction of mis-registrations hence yielding mislabels for the classification task in this work. Mislabeling occurs when a field-surveyed coniferous tree stem is assigned to a segmented deciduous tree crown or vice versa. In the semi-supervised learning literature, a number of studies trained learning models that are robust to such noise by modifying the learning model to explicitly account for the noise (Mnih and Hinton 2012; Natarajan et al. 2013; Reed et al. 2014), although these studies did not necessarily correct mislabels for external use. Other studies attempted to eliminate/correct mislabels by training learning models and identified mislabels by performing statistical inference on the classification result of the trained models (Bhadra and Hein 2015; Brodley and Friedl 1999). These studies either used a small noise-free dataset or, when that was not possible, made assumptions about the tolerable amount of noise in their data to train their learning models for identifying mislabels. For the latter scenario, some studies reported successful identification of mislabels in the presence of up to 40% noise in the training data (Brodley and Friedl 1999). Unlike general RGB images that are specifically designed for human visual comprehension, remotely sensed LiDAR-represented tree crowns are difficult and uncertain for human experts to classify, making it infeasible to create a noise-free dataset to start with. Therefore, we performed mislabel correction through ensemble filtering (Brodley and Friedl 1996), which is derived by a series of resampling and statistical inferences.

We built 100 4×2D-input networks, and each network was trained using a random sample of 80 deciduous and 80 conifer instances from our labeled dataset. Random sampling was performed without replacement: once all corresponding labeled instances were used, we started over and continued until all 100 networks were built. This randomization pattern ensured that all instances of a class had (almost) equal contributions across all of the networks in the training process. To train the networks, we used the Keras deep learning library: we set the loss function to categorical cross entropy and ran the Adam optimizer (learning rate = 0.01) (Kingma and Ba 2014). The training of each network was performed for three epochs in order to ensure that the process converged to a reasonable state, i.e., the training accuracy was lifted from the base accuracy of 50% but did not reach an overfitting phase.

For each network *n*, we computed the average of the test accuracies of *n* over the 180 augmented forms ($acc_{ni}$) for every instance *i* in the labeled dataset if *i* was not used in training *n*. Assuming instance *i* is correctly labeled, its test accuracy should on average be equal to the training accuracy of the trained network *n* ($acc_n$). On the other hand, when instance *i* is mislabeled, its test accuracy should on average be equal to the symmetric value of the training accuracy of *n* about the base accuracy of 50% ($1 - acc_n$). Therefore, if $acc_{ni}$ is less than the symmetric value of the training accuracy of *n* about 50%, i.e., $acc_{ni} < 1 - acc_n$, it is very likely that *i* is mislabeled. Using all 100 networks, we generated values of $acc_{ni} - (1-acc_n)$ per each instance *i* and used these values to perform a T-test on whether their mean was less than zero. If the T-test indicated that an instance was mislabeled, we flipped the label for that instance. We repeated the process of training 100 networks, performing T-tests, and flipping mislabels until no mislabels were identified. Since 2,528 T-tests were performed in each iteration, we used the significance level of $10^{-8}$ for the T-tests. This significance level, according to the conservative Bonferroni principle, would not allow a false positive rate of more than 0.0025% per iteration.

## 2.5 Classification and evaluation

After correcting potential mislabels, we used an ensemble of 50 networks to perform the classification. We trained each network on a random sample of 100 deciduous and 100 coniferous instances using the Adam optimizer with a learning rate of 0.01. Random sampling was performed without replacement as mentioned above. The ensemble training scheme was used to minimize the bias of unbalanced training instances in each class, i.e., to train each network on a balanced sample while taking advantage of the entire dataset. To perform the classification for a given instance, we averaged over the softmax probabilities produced by all of networks that did not use that instance for training and assigned the class as that with the larger average probability. This training and testing pattern allowed us to produce cross validated classification accuracies for all of the instances in our dataset. We performed the same ensemble procedure for both the DSM×4 and the 4×2D formats. The training was run for fifteen epochs for every DSM×4 input network, but five epochs appeared to be sufficient for every 4×2D input network. For the rest of experiments, we used the 4×2D format because of the lower computational load.

To investigate the effect of the training data size, we created stratified random subsamples of our dataset. We subsampled 20%, 40%, ..., 100% of the deciduous and coniferous trees and performed the cross validated classification procedure described above for each subsampled dataset. We adjusted the size of resampling instances in proportion to the subsample size, though the number of ensemble networks was held constant. To quantify the effect of data augmentation, we measured the cross validation accuracies for when 20, 40, ..., 180, 240, 300, and 360 rotations of each instance were included. We then looked into the effects of the domain parameters: we ran the cross validation experiment excluding leaf-off data, excluding leaf-on data, using non-normalized intensities for leaf-on data, and excluding intensity values (using binary values representing existance of a point per pixel). When excluding leaf-on and leaf-off data, we decreased the size of the last two dense layers before the softmax layer to 16 and 8 units respectively to account for the smaller input size. We also inspected the correlation between the point density of a crown cloud and the probability of the softmax output unit associated with the correct label of the crown cloud to determine how point density affected the classification accuracy. Lastly, we stratified the classification result to overstory (dominant and co-dominant) and understory (intermediate and overtopped) trees to inspect how crown class affected the classification performance.

## 3 Results and Discussion

### 3.1 Mislabel correction

The process of mislabel correction converged after 13 iterations and increased the number of conifers from 124 to 214 and decreased the number of deciduous trees from 2404 to 2314 (Figure 1-a). According to the original field measurements (Table 1), 7.27% of the trees in RF are conifers, which is slightly lower than the result after correcting mislabels – 8.46% conifers. The reason for this slight difference may be the relative difficulty in segmenting deciduous trees compared to coniferous trees due to the variety of crown shapes and the looser, interwoven foliage, which creates complicated, difficult-to-distinguish LiDAR point patterns (Vauhkonen et al. 2012). This effect likely resulted in larger rate of undetected deciduous trees after segmentation and registration with the field data. In total, the labels for 35 of the initial 124 (28.22%) conifers and 125 of 2404 (5.20%) initial deciduous trees were flipped. These unbalanced flip rates concur with the dominant presence of deciduous trees, i.e., if a field deciduous tree is mis-registered to a LiDAR crown, the crown is likely another deciduous tree (yielding no mislabel) while for a mis-registered field conifer this is not the case. Over the 13 iterations of the mislabel correction procedure, the average training accuracy of the 100 networks started at 67.1% and plateaued at 83.6% (Figure 1-b). This trend suggests that a number of highly likely (controlled by the T-tests) mislabels were corrected, improving the model accuracy, while less likely mislabels were left unchanged, resulting in the accuracy plateau and prohibition of overfitting. Overall, the mislabel correction process produced more realistic labels by increasing the number of coniferous trees from 4.90% to 8.46% within the 2528 segmented tree crowns.

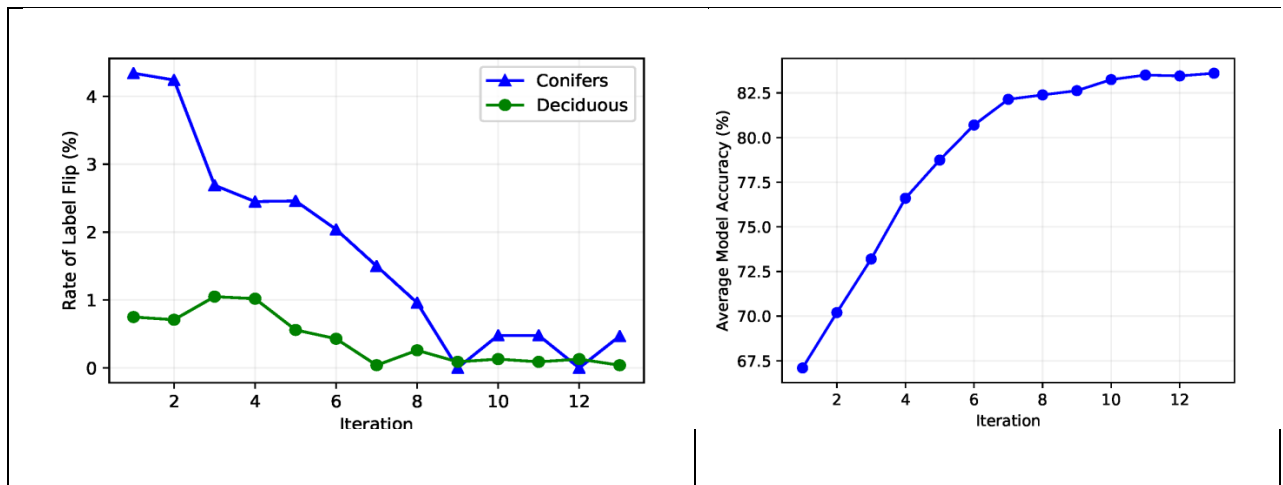

Figure 1. (a) Rates of flip for coniferous and deciduous trees over the 13 iterations of the mislable correction process; and (b) average training accuracy of 100 networks over the 13 iterations.

## 3.2 Classification accuracy

The cross-validated accuracies associated with the DSM×4 representation were 80.4±5.3% for conifers and 90.1±1.3% for deciduous trees at a confidence level of 95%. The equivalent classification accuracies associated with the 4×2D representation were 82.7±5.1% and 90.2±1.3%, respectively for coniferous and deciduous trees (Figure 2). Higher accuracy values associated with the 4x2D representation were insignificant and are likely due to the fact that this format was used for the mislabel correction process, which might have slightly biased the data. As mentioned, the DSM×4 format more closely resembles 3D data, which together with the richer early-fused network, have the potential to achieve higher classification accuracies. However, the 4×2D format with a late-fused network could achieve similar accuracies while incurring less computational load.

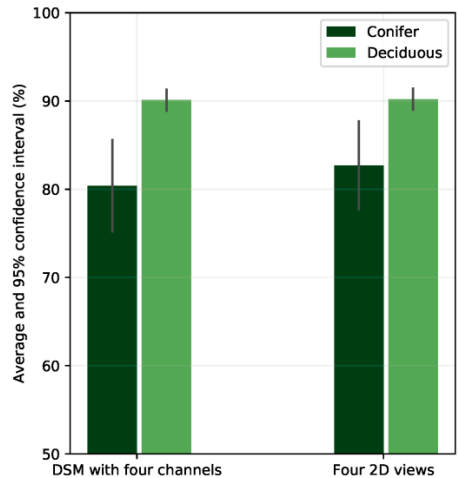

Figure 2. Classification accuracy when using the two representational formats derived from discretization of LiDAR point clouds.

## 3.3 Effect of training data size on the classification

Increasing the size of training data improved the classification accuracies. For deciduous trees the accuracy plateaued when using only 40% of the original dataset (~925 deciduous and ~86 coniferous trees) but for coniferous trees, the accuracy appeared to be increasing with even more number of conifer instances than the original 214 ones (Figure 3). This observation suggests that a balanced dataset of close to one thousand instances per class would likely have been an optimal dataset for this classification task and could have brought the accuracy of coniferous trees closer to that of deciduous trees.

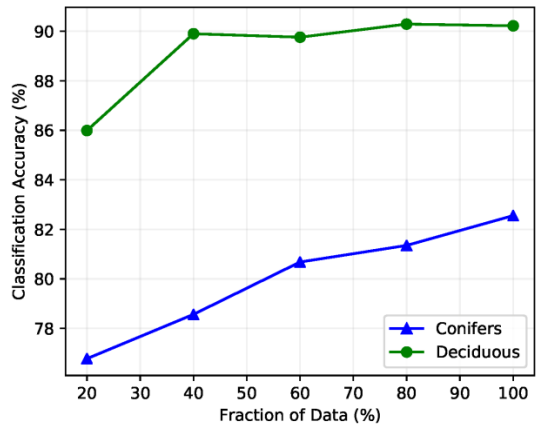

Figure 3. Classification accuracies measured against the size of the training data. Each symbol in the diagram represents the average of 20 observations.

## 3.4 Effect of data augmentation on classification

Including a greater number of rotational augmentations per instance slightly improved the classification accuracies. Using only 20 rotations per instance resulted in 73.8% accuracy for coniferous trees and 87.7% accuracy for deciduous trees, which are lower than when using the original 180 rotations. The improvement in classification plateaued at ~60 rotations for deciduous trees and ~150 rotations for coniferous trees (Figure 4). Having more deciduous trees likely resulted in a smaller number of rotations/augmentations to be sufficient for the classification task. Although a higher number of rotations could compensate for the small number of coniferous training instances to some extent, augmentations are unlikely to match the classification quality provided by a higher number of real training instances.

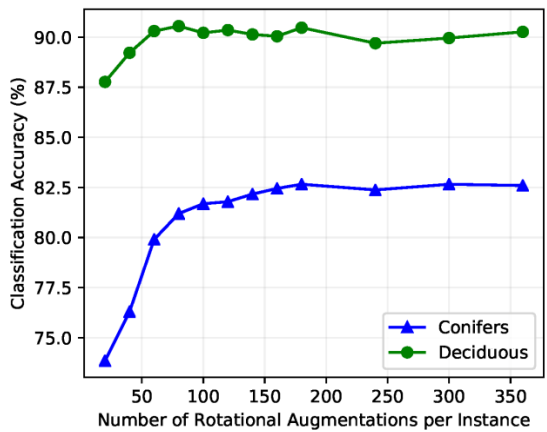

Figure 4. Classification accuracies measured against the number of rotational augmentations per instance.

## 3.5 Effect of domain data on classification

Excluding the leaf-off data resulted in a remarkable decrease in classification accuracy for the conifers (from 82.7% to 61.2%) and a minor decrease in accuracy for deciduous trees (from 90.5% to 89.6%), while excluding the leaf-on data resulted in a minor decrease in accuracy for conifers (from 82.7% to 81.6%) and a negligible increase in accuracy for deciduous trees (from 90.5% to 90.7%) (Figure 5). This observation indicates that, despite the much lower point density, the leaf-off data provided the most useful features for the classification task, which concurs with the result of the previous work (Kim et al. 2011; Reitberger et al. 2008). As conifers do not lose their dense foliage during the winter, the leaf-off LiDAR points could represent their crown shapes even at a low density while the deciduous trees may only be represented by a few random LiDAR points returning from their defoliated branches. The dense leaf-on data could on the other hand represent the crown shapes for both conifers and deciduous trees and was used here for segmentation of the individual tree crowns. Attempting to distinguish the crown shapes of deciduous and coniferous trees using leaf-on data is likely less efficient than distinguishing between a random point pattern (a deciduous tree) and a crown-like shape (a coniferous tree) using the leaf-off data. However, for identifying species, which is a more complicated classification task and a subject of future work, the high density leaf-on data may be more useful.

Using binary values instead of the intensity values resulted in a remarkable decrease in classification accuracy for conifers (from 82.7% to 69.2%) and only a negligible increase in accuracy for deciduous trees (from 91.5% to 92.1%) (Figure 5). Using the normalized intensity values for the leaf-on data (when excluding the leaf-off data) compared with using non-normalized values, seemed to make minor, insignificant improvements in the classification accuracies for conifers (from 60.3% to 61.2%) and deciduous trees (from 88.9% to 89.6%) (Figure 5). Although LiDAR intensity values were useful for the classification, the uncontrollable atmospheric factors present during LiDAR acquisition and the discretization to an eight-bit format likely introduced some level of noise, yielding the process of intensity normalization less effective than expected.

Lastly, some domain data, i.e., the leaf-off data and the intensity values, appeared to be very important in the classification task, which is evident in the remarkable changes in the accuracy for conifers (Figure 5). However, as observed by only slight changes in the accuracy of deciduous trees, abundance of training data likely compensates for the absence of a subset of important domain data.

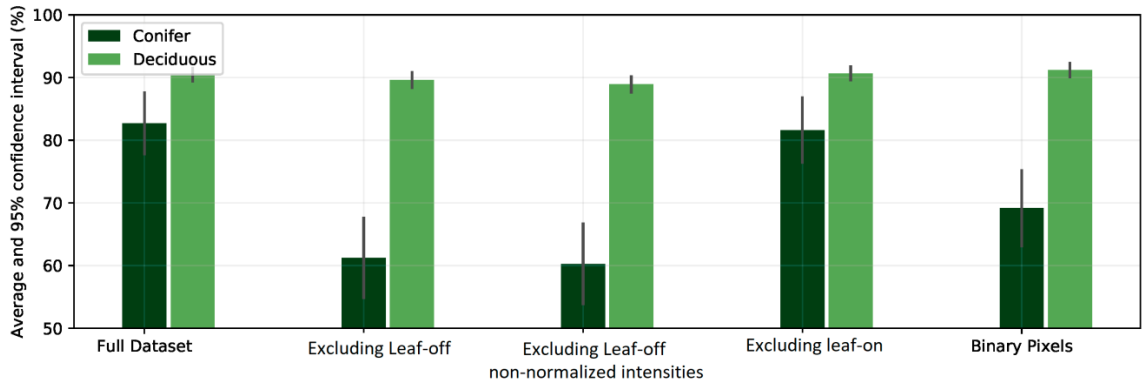

Figure 5. Classification accuracy when excluding domain data.

### 3.6 Effects of crown class and point density on classification

For overstory trees, the cross validated classification accuracy was 92.1±4.7% for conifers and 87.2±2.2% for deciduous trees. The classification accuracy for understory trees was 69.0±9.8% for conifers and 92.1±1.4% for deciduous trees (Figure 6). The crown of an understory tree is typically captured only partially by airborne LiDAR, as it is covered by the overstory trees. The partial shapes of these crowns decrease the classification power, likely yielding the correlated accuracies to become easily biased by the abundance of deciduous instances compared with coniferous instances. In contrast, the crowns of overstory trees are captured more completely allowing for a more powerful classification process. Lastly, we could not identify any significant correlation between point density (neither leaf-off nor leaf-on) and the classification accuracy (neither for overstory nor for understory trees). This observation does not concur with previous work reporting a positive correlation between accuracy and point density (Li et al. 2013). The reason is likely that the classification task is primarily driven by the leaf-off data, the point density range of which is too small (0.1-6.0 pt/m² for the middle 95%) to surface any effect. Moreover, the partial crowns captured may feature high point densities but are not easy to classify due to their incomplete shapes.

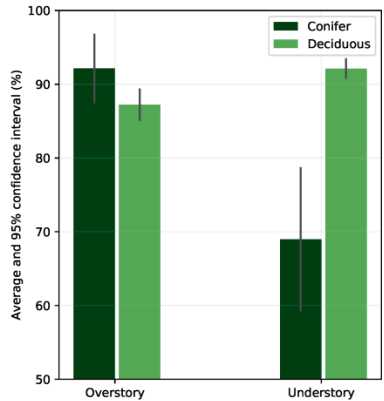

Figure 6. Classification accuracy of overstory and understory trees.

# 4 Conclusions

Airborne LiDAR point clouds representing individual trees can be used to predict tree attributes such as tree type. Previous work exploited shallow learning techniques that require the engineering of useful features by a human expert. In this work, we used deep learning CNNs to classify crown point clouds as coniferous or deciduous trees. We segmented individual trees from the LiDAR point clouds and registered them with field-surveyed trees to create training data. We designed two different discrete representations of a crown's 3D point cloud and the corresponding deep learning architectures. We used ensemble learning schemes including several networks trained on balanced subsamples of training data to perform mislabel correction (driven by statistical tests) and to measure the cross-validated classification accuracies in different scenarios.

Our investigation of the coniferous/deciduous deep learning classification showed that a set of 2D views/profiles of a 3D point cloud are not only more efficient to be processed but also can yield similar or even higher accuracies compared with bulkier 2.5D (or even 3D) representations. Moreover, late fusion of features in a CNN architecture may provide equivalent performance as compared with an early-fused architecture while incurring less computational load. Although data augmentation can help improve the classification accuracy, a higher number of real training instances can provide much stronger improvements. As we observed, leaf-off LiDAR data, despite its much lower point density in comparison with the leaf-on data, was the main source of useful information, which is likely associated with the perennial nature of conifer foliage. LiDAR intensity values also proved to be useful for the classification, although we could not obtain a significant improvement by normalizing the intensity values. A large number of training instances may compensate for the lack of a subset of important domain data. Lastly, we observed much higher and balanced classification accuracies for overstory trees (~90%) as compared with understory trees (~90% for deciduous and ~65% for coniferous), which is likely associated with capturing only partial shapes of many understory tree crowns using airborne LiDAR.

The results presented indicate that deep learning can effectively and efficiently be used for classifying tree type based on airborne LiDAR point clouds representing individual tree crowns, which is a step forward to operational tree-level remote quantification of large-scale forests. Although further experiments using richer datasets and for more complicated prediction tasks (e.g., species classification) are required, deep learning provides the feasibility of automatic extraction of optimal features toward the prediction task. This unique deep learning characteristic brings about the potentials for successful prediction tasks in different domains such as remote sensing and biomedical image analysis, where the data modalities are not friendly to the human perceptual system and have likely operated using suboptimal human-designed features.